\documentclass[a4paper,fleqn]{cas-sc}

\usepackage[numbers]{natbib}

\usepackage[scaled=.90]{helvet}
\AtBeginEnvironment{figure}{\renewcommand{\sfdefault}{cmss}}
\AtBeginEnvironment{table}{\renewcommand{\sfdefault}{cmss}}
{\fontencoding{T1}\fontfamily{cmss}\selectfont}
\DeclareFontShape{T1}{cmss}{b}{n}{<->ssub * cmss/bx/n}{}
\makeatletter
\newcommand\cas@stocksans{\renewcommand{\sfdefault}{cmss}}
\let\cas@head@orig\@oddhead
\let\cas@foot@orig\@oddfoot
\renewcommand\ps@cas{%
  \def\@oddhead{{\cas@stocksans\cas@head@orig}}%
  \def\@oddfoot{{\cas@stocksans\cas@foot@orig}}%
  \let\@evenhead\@oddhead
  \let\@evenfoot\@oddfoot}
\ps@cas
\makeatother

\newcommand{\tabgroup}[2]{\addlinespace[1.5pt]%
  \multicolumn{#1}{@{}l}{\itshape #2}\\[1.5pt]}
\newcommand{\good}[1]{\textcolor{okline}{\textbf{#1}}}
\newcommand{\bad}[1]{\textcolor{badline}{#1}}

\usepackage{placeins}   
\usepackage[most]{tcolorbox}
\usepackage{listings}
\usepackage{upquote}

\definecolor{okline}{HTML}{2E7D46}
\definecolor{okfill}{HTML}{EDF6EF}
\definecolor{badline}{HTML}{B23B3B}
\definecolor{badfill}{HTML}{FBEDEB}
\newtcolorbox{qacorrect}[1]{colback=okfill,colframe=okline,coltitle=white,
  fonttitle=\sffamily\bfseries\footnotesize,title={#1},boxrule=0.6pt,arc=2pt,
  fontupper=\footnotesize,left=6pt,right=6pt,top=3pt,bottom=3pt,
  before skip=8pt,after skip=8pt}
\newtcolorbox{qawrong}[1]{colback=badfill,colframe=badline,coltitle=white,
  fonttitle=\sffamily\bfseries\footnotesize,title={#1},boxrule=0.6pt,arc=2pt,
  fontupper=\footnotesize,left=6pt,right=6pt,top=3pt,bottom=3pt,
  before skip=8pt,after skip=8pt}
\definecolor{promptline}{HTML}{3E5C76}
\definecolor{promptfill}{HTML}{F3F6F9}
\definecolor{plusline}{HTML}{C8743F}
\tcbset{agentcard/.style={enhanced,colback=white,colframe=#1,colbacktitle=#1,coltitle=white,
  fonttitle=\sffamily\bfseries\footnotesize,fontupper=\fontsize{7.5}{9}\selectfont,fontlower=\fontsize{7.5}{9}\selectfont,
  boxrule=0.6pt,arc=2pt,left=4pt,right=4pt,top=2pt,bottom=2pt,toptitle=1pt,bottomtitle=1pt,
  segmentation style={draw=#1!45,dashed,line width=0.4pt},
  halign upper=flush left,halign lower=flush left,space to lower,valign lower=top,middle=1.5pt}}

\definecolor{tagrole}{HTML}{3E5C76}
\definecolor{taginput}{HTML}{6F8FAF}
\definecolor{tagoutput}{HTML}{2E7D46}
\newcommand{\cardtag}[2]{\tcbox[on line,colback=#1,colframe=#1,boxrule=0pt,arc=1.5pt,
  boxsep=0pt,left=0pt,right=0pt,top=0.8pt,bottom=0.8pt,
  fontupper=\sffamily\bfseries\scriptsize\color{white}]{\makebox[1.05cm][c]{#2}}}
\newcommand{\cardrow}[3]{\par\noindent\hangindent=1.25cm\hangafter=1
  \cardtag{#1}{#2}\hspace{0.2cm}#3\par}
\newcommand{\qalabel}[1]{\smallskip\noindent{\sffamily\bfseries\footnotesize #1}\ }

\definecolor{sqlkw}{HTML}{15467E}
\definecolor{sqlstr}{HTML}{9A5B00}
\definecolor{sqlcomment}{HTML}{7A7A7A}
\definecolor{sqlrule}{HTML}{BFBFBF}
\lstdefinestyle{crashsql}{
  language=SQL, sensitive=false, upquote=true,
  basicstyle=\ttfamily\footnotesize,
  keywordstyle=\color{sqlkw}\bfseries,
  stringstyle=\color{sqlstr},
  commentstyle=\color{sqlcomment}\itshape,
  showstringspaces=false,
  breaklines=true, breakatwhitespace=true,
  postbreak=\mbox{\textcolor{sqlrule}{$\hookrightarrow$}\space},
  columns=fullflexible, keepspaces=true,
  aboveskip=3pt, belowskip=2pt,
  backgroundcolor=\color{white},
  frame=l, framerule=1.2pt, rulecolor=\color{sqlrule},
  framexleftmargin=5pt, xleftmargin=7pt, xrightmargin=2pt,
  literate={>}{{\textcolor{sqlkw}{\textgreater}}}1,
  morekeywords={EXTRACT,MONTH,FETCH,FIRST,ROWS,ONLY}
}

\newcommand{\ind}{\mathbf{1}}
\newcommand{\exec}{\operatorname{exec}}
\newcommand{\code}[1]{\texttt{\footnotesize #1}}
\newcommand{\Db}{\mathcal{D}}
\newcommand{\Sch}{\mathcal{S}}
\providecommand{\K}{}\renewcommand{\K}{\mathcal{K}}
\newcommand{\Fsys}{F}
\newcommand{\rows}{\operatorname{rows}}

\newlength{\heatw}
\newcommand{\heat}[1]{%
  \pgfmathtruncatemacro{\heatint}{round(#1)}%
  \ifnum\heatint<50
    \edef\hc{badline!\the\numexpr(50-\heatint)*6/10\relax!white}%
  \else
    \edef\hc{okline!\the\numexpr(\heatint-50)*6/10\relax!white}%
  \fi
  {\setlength{\fboxsep}{1.5pt}\colorbox{\hc}{\makebox[\heatw][r]{\strut #1}}}}

\begin{document}
\let\WriteBookmarks\relax

\shorttitle{CrashDB-QA: Schema-Grounded Agentic Workflow for Crash Data}
\shortauthors{J.~Zhu et~al.}

\title[mode = title]{
From Questions to Queries to Insights: A Schema-Grounded Agentic Workflow for Natural-Language Crash Data Analysis}

\author[1]{Jingwen Zhu}
\credit{Conceptualization, Methodology, Software, Data curation, Investigation, Writing -- original draft}

\author[1]{Keshu Wu}
\cormark[1]
\ead{kwu84@wisc.edu}
\credit{Conceptualization, Methodology, Formal analysis, Writing -- original draft}

\author[1]{Steven T. Parker}
\credit{Conceptualization, Data curation, Resources, Writing -- review \& editing}

\author[1]{Bin Ran}
\credit{Supervision, Writing -- review \& editing}

\author[1]{David A. Noyce}
\credit{Supervision, Writing -- review \& editing}

\affiliation[1]{organization={Department of Civil and Environmental Engineering, University of Wisconsin--Madison},
    city={Madison},
    state={WI},
    country={United States}}

\cortext[1]{Corresponding author.}

\begin{abstract}
Transportation agencies maintain extensive crash databases to identify safety priorities, evaluate countermeasures, and guide resource allocation. Making these data accessible to practitioners requires flexible ways to translate analytical questions into reliable results. Existing dashboards support predefined analyses, while additional requests often depend on specialists who understand database queries, coded fields, and agency conventions. This study presents CrashDB-QA, a fully local, schema-grounded agentic workflow that enables natural-language analysis of crash databases. Specialized components interpret questions, generate and validate read-only SQL, execute queries, and explain the results. A shared knowledge base grounds query construction in the agency's data dictionary and safety-analysis conventions, while recorded intermediate outputs support inspection and auditing. An optional extension, CrashDB-QA+, checks the alignment among the question, query, and execution result and supplies evidence for iterative repair. We evaluate the workflow on a frozen snapshot of 4.4 million Wisconsin crash records using a new benchmark of 300 questions across ten analytical categories and nine local open-weight models. Domain grounding substantially improves query accuracy, and a code-specialized 7B model provides the best accuracy-latency balance among the tested models. Compositional analyses remain more difficult than routine retrieval, particularly comparisons between groups or periods. Verification improves performance when the Judge uses authoritative field definitions and the Refiner receives execution evidence, whereas additional checking alone offers no consistent benefit. These findings demonstrate the potential of locally deployed, domain-grounded workflows to broaden access to crash analysis and identify the reasoning and verification capabilities needed for reliable use.
\end{abstract}



\maketitle

\section{Introduction}\label{sec:intro}
Transportation agencies collect millions of crash records to support data-driven safety management. These records provide a foundation for comparing crash patterns across locations and periods, identifying high-risk locations, and examining the safety of vulnerable road users. Translating this evidence into actionable insights helps agencies prioritize engineering improvements, evaluate countermeasures, and allocate enforcement resources. Their importance is underscored by the persistent burden of road traffic deaths and serious injuries worldwide \citep{who2023road}. Crash databases provide the evidence needed to identify high-risk locations \citep{hauer2002eb}, evaluate countermeasures, and inform safety policies, but this evidence becomes useful only when practitioners can retrieve and interpret it. Dashboards and visualization platforms have improved access through maps, summaries, and predefined filters \citep{communitymaps2024}, while predictive models have extended analysis toward anticipating crash hotspots \citep{zhu2026forecasting}. However, questions beyond these predefined functions often require someone to identify the relevant fields, interpret coded values, and write Structured Query Language (SQL). The resulting dependence on technical specialists limits how readily practitioners can explore the data and pursue follow-up questions.

Large language models (LLMs) offer a promising way to bridge this gap by allowing practitioners to express analytical requests in natural language. Their emerging transportation applications include signal control \citep{lai2025llmlight}, spatiotemporal forecasting \citep{li2024urbangpt}, tool-augmented traffic analysis \citep{zhang2024trafficgpt,da2024openti}, and connected-vehicle data interpretation \citep{wu2025v2x}, as reviewed by \citet{nie2025roles}. In traffic safety, LLMs have been explored for research assistance \citep{zheng2023chatgpt}, crash-narrative analysis \citep{mumtarin2023crash,bharati2026benchmarking}, contributing-factor identification \citep{abdelrahman2025causation}, severity inference \citep{zhen2024severity,yang2026transportagents}, crash-record modeling \citep{fan2024crashllm}, risk assessment \citep{zhao2025safetraffic}, and personal-information screening \citep{ma2026agentic}. Studies with state crash data have also used LLMs to correct the coding of secondary crashes \citep{zhang2026kentuckyllm} and to link crash narratives with pavement-condition records \citep{gottimukkala2026pavement}. A University Transportation Center prototype has enabled natural-language queries and chart generation over a statewide crash repository \citep{tang2024neutcllm}. Its report presents demonstrations rather than a systematic evaluation of answer accuracy. These applications demonstrate the potential of language-based interaction with safety information. Answering an analyst's question directly from a structured crash database, however, requires a further capability: translating the request into a computation that follows the agency's definitions and returns the intended result.

Text-to-SQL research provides a foundation for this capability \citep{katsogiannis2023survey}. Neural schema-aware models \citep{zhong2017seq2sql,wang2020ratsql,guo2019irnet} and more recent LLM-based approaches \citep{rajkumar2022evaluating,gao2024text2sql} have advanced natural-language querying, supported by benchmarks such as Spider and BIRD \citep{yu2018spider,li2023bird} and methods for decomposition, schema linking, collaboration, and correction \citep{pourreza2023dinsql,talaei2024chess,wang2025macsql}. In crash analysis, however, matching words to columns is only part of the problem. Consider a request to identify counties with the greatest severity-weighted crash burden. A query that counts crashes may execute successfully and return a convincing ranking, yet omit the weighting that defines the requested measure \citep{aashto2010hsm,elvik2009handbook}. Likewise, a seasonal analysis may use the correct records but group them by month, and a two-period comparison may inadvertently apply the same condition to both groups. These \emph{executable-but-wrong} queries expose the central challenge: successful execution does not establish that an answer is analytically correct. The system must connect the analyst's intent to field meanings, coded values, and the rules used to construct the requested quantity.

This challenge links domain grounding, verification, and deployment. A governed crash database has an authoritative data dictionary, but abbreviated column names alone may not reveal its definitions or coding conventions. Query generation must therefore use documented knowledge rather than rely on the model's prior assumptions. Verification must then examine whether the generated query and its returned result satisfy the request. Reasoning and tool use \citep{yao2023react,wei2022cot}, consistency across candidates \citep{wang2023selfconsistency}, iterative refinement \citep{madaan2023selfrefine,shinn2023reflexion}, and model-based evaluation \citep{zheng2023judge} offer useful mechanisms, but their presence does not guarantee correctness. Self-correction without external feedback can be unreliable \citep{huang2024selfcorrect}, and evaluation in traffic safety has also revealed differences between predictive accuracy and expert agreement \citep{bhagat2025accuracy}. A database workflow consequently needs checks grounded in both authoritative definitions and execution evidence. At the same time, sensitive or restricted crash records may preclude externally hosted models, making local inference, read-only access, and an inspectable question-to-answer process practical requirements. Together, these needs motivate a workflow in which language-based reasoning is supported by domain knowledge and constrained by deterministic database controls.

We present \emph{CrashDB-QA}, a fully local, schema-grounded agentic workflow designed around these requirements. Specialized components interpret the question, link it to the schema, generate SQL, validate and execute the query, and explain the result. A shared knowledge base combines the agency's data dictionary, stored value formats, common synonyms, and safety-analysis conventions to support retrieval-grounded generation \citep{lewis2020rag}. Deterministic controls enforce read-only access and schema validity, while recorded intermediate outputs make each answer traceable. The optional \emph{CrashDB-QA+} extension adds a Judge that checks the alignment among the question, query, and execution result, together with a Refiner that supplies feedback for revision. We evaluate the workflow on a frozen snapshot of $4.4$ million Wisconsin crash records maintained through the Wisconsin Transportation Portal, developed by the University of Wisconsin--Madison Traffic Operations and Safety Laboratory in collaboration with the Wisconsin Department of Transportation. A new $300$-question benchmark and nine local open-weight models allow us to examine accuracy, latency, compositional difficulty, and the effects of verification. The results show that domain grounding makes compact local models substantially more effective, while comparisons between groups or periods remain a persistent weakness. They also show that verification improves performance only under suitable designs: grounding the Judge and supplying execution evidence to the Refiner produces gains that additional checking alone does not.

This paper makes two main contributions. First, we introduce \textbf{CrashDB-QA}, an auditable workflow that integrates domain-grounded natural-language querying, deterministic database controls, and result explanation, with an optional execution-informed verification and repair loop. Second, we introduce an \textbf{execution-based benchmark and evaluation} comprising $300$ questions across ten analytical categories in retrieval and reasoning tiers. The evaluation identifies accuracy--latency tradeoffs among local models, distinguishes routine retrieval from difficult analytical composition, and establishes the benefits and limitations of grounding and iterative repair.

\section{The CrashDB-QA Framework}\label{sec:method}

\subsection{Problem formulation}\label{sec:problem}

CrashDB-QA enables transportation safety analysts to obtain answers from an agency's crash database through natural-language questions. Let $\Db$ denote the database and $\Sch$ its column schema. An authoritative data dictionary specifies the meaning and coding of each field. Given a question $q$, the system produces a SQL query $Q$, its execution result $R$, and a plain-language explanation $a$:
\begin{equation}
\Fsys:\ q\ \longmapsto\ (Q,\,R,\,a),
\qquad R=\exec(Q,\Db).
\label{eq:task}
\end{equation}
The workflow operates entirely on agency-controlled hardware and admits only queries that satisfy the database access constraints. We denote this admissible query set by $\mathcal{V}$: each query must be read-only, reference existing schema elements, and satisfy the implemented SQL validation checks.

We evaluate answer correctness by comparing the returned result with the result of an analyst-written reference query $Q^\star$. This execution-based criterion accommodates different SQL formulations that produce equivalent answers. Let $\rows(R)$ denote the normalized multiset of result rows, with column names, column order, and specified formatting differences removed before comparison. For $R^\star=\exec(Q^\star,\Db)$, execution match is
\begin{equation}
\mathrm{EM}(R,R^\star)
=
\ind\!\big[\rows(R)=\rows(R^\star)\big].
\label{eq:em}
\end{equation}
The design objective is therefore to maximize execution accuracy while maintaining admissible database access:
\begin{equation}
\max_{\Fsys}\ 
\mathbb{E}_{q}\!\left[
\mathrm{EM}(R_q,R_q^\star)
\right]
\qquad
\text{subject to}\qquad
Q_q\in\mathcal{V}
\quad\text{for every executed query }Q_q.
\label{eq:objective}
\end{equation}
Local deployment is an additional architectural requirement: model inference, query execution, and result processing remain within the agency's environment.

The central difficulty is that an admissible query can still compute the wrong quantity. Consider a request to identify the five counties with the greatest severity-weighted crash burden in a specified year. A query that ranks counties by \code{COUNT(*)} may use valid columns and execute successfully, while omitting the severity weights that define the requested measure. We refer to such outputs as \emph{executable-but-wrong} queries. Addressing them requires both accurate grounding before execution and an assessment of question--query--result alignment afterward. Reference results support evaluation, but are unavailable during deployment; operational verification must therefore estimate correctness from the available evidence.

\subsection{Framework overview}\label{sec:overview}

CrashDB-QA organizes natural-language analysis as a sequence of specialized components with explicit inputs and outputs. Its six-component core comprises the \emph{Interpreter}, \emph{Schema Linker}, \emph{Generator}, \emph{Validator}, \emph{Executor}, and \emph{Explainer}. The Interpreter and Schema Linker establish the analytical intent and relevant fields; the Generator and Validator construct an admissible query; and the Executor and Explainer produce the result and its explanation. A shared knowledge base supplies authoritative field definitions and analysis conventions at the stages where they are needed. The optional extension, \emph{CrashDB-QA+}, inserts a \emph{Judge} and a \emph{Refiner} between execution and explanation to assess the result and guide revision.

\begin{figure}[pos=t]
\centering
\includegraphics[width=\linewidth]{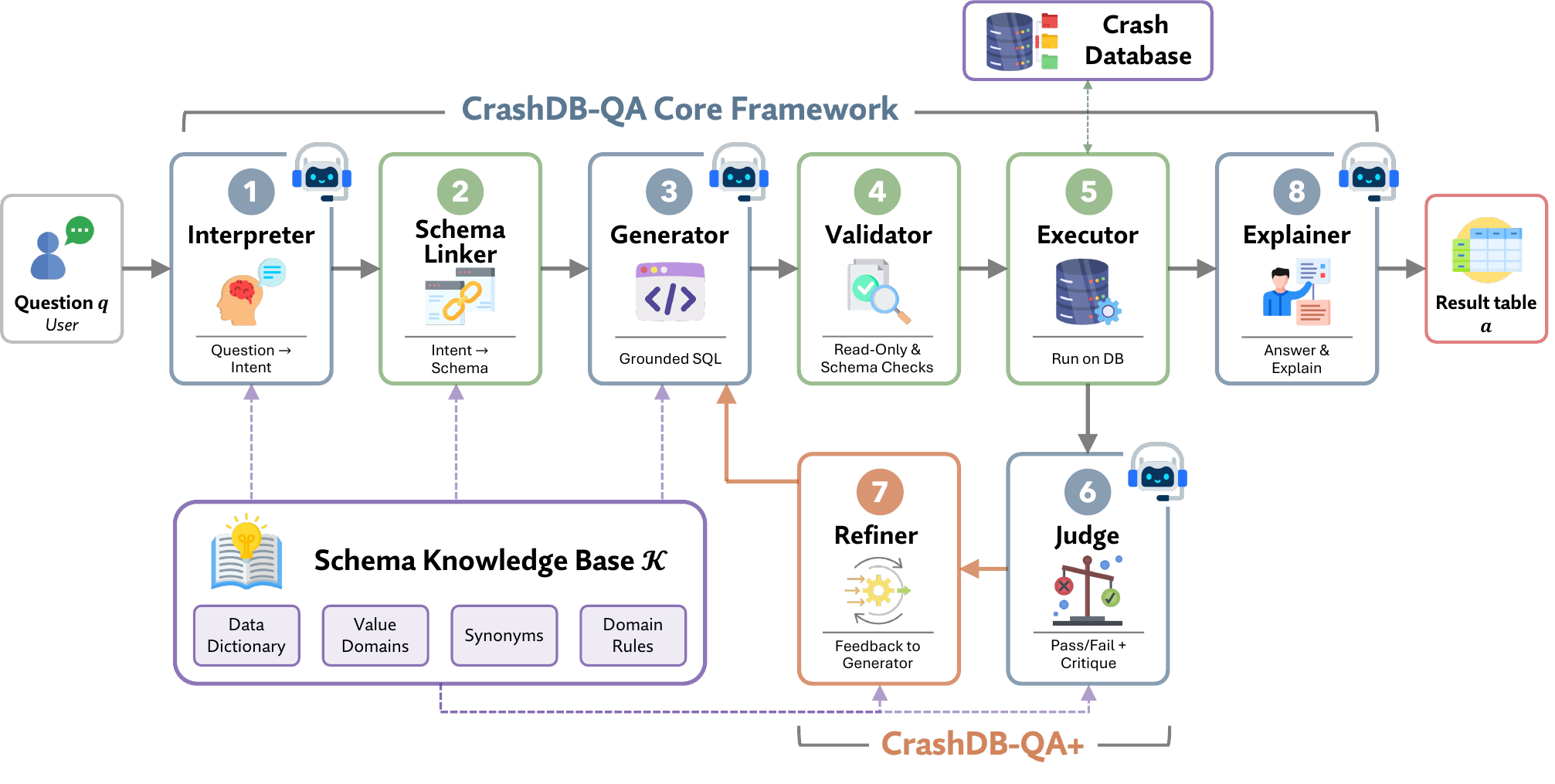}
\caption{CrashDB-QA architecture. The six-component core connects question interpretation, grounded query construction, controlled execution, and result explanation. The optional CrashDB-QA+ extension, shown in orange, adds verification and iterative repair.}
\label{fig:arch}
\end{figure}

The database execution boundary provides the organizing structure in Figure~\ref{fig:arch}. Steps~1--4 operate on the question, schema knowledge, and generated SQL before a query accesses crash records. Step~5 executes an admitted query, making its returned values and result structure available for subsequent assessment. In the core, this result proceeds directly to explanation. In CrashDB-QA+, Steps~6--7 examine the question, query, and execution evidence and return feedback to query construction when a check fails. Step~8 explains the selected result. This arrangement allows errors to be traced to interpretation, schema selection, query construction, execution, or explanation rather than remaining hidden within a single model response.

Language models perform the tasks that require language interpretation: intent extraction, SQL generation, result assessment, and explanation. Deterministic components perform schema retrieval, query validation, database execution, and feedback assembly. The execution order and retry limits of all components are controlled by code, with the Judge's verdict determining whether semantic refinement continues. CrashDB-QA is thus an \emph{agentic workflow} with predefined orchestration. This structure supports predictable database access and a recorded trace of intermediate outputs, while retaining model flexibility within individual tasks.

\subsection{Governed-schema knowledge base}\label{sec:rag}

The knowledge base $\K$ connects the language used by analysts to the database's documented meanings and stored representations. This connection is important for the crash-level table, whose $169$ columns include abbreviated identifiers such as \code{INJSVR} and \code{CONSZONE}. Column names provide limited guidance about their interpretation, permissible values, or use in an analytical measure. Built before deployment and retrieved during analysis, $\K$ supplies this information through schema-focused retrieval-augmented generation \citep{lewis2020rag}. It contains column-level knowledge and analytical conventions:
\begin{equation}
\begin{aligned}
\K &= \big(\K_{\mathrm{col}},\,\K_{\mathrm{conv}}\big),\\
\K_{\mathrm{col}}
&= \big\{(c,D_c,V_c,W_c):c\in\Sch\big\},\\
\K_{\mathrm{conv}}
&= \big\{(p,\sigma_p):p\in\mathcal{P}\big\}.
\end{aligned}
\label{eq:kb}
\end{equation}
For each column $c$, $D_c$ is its dictionary description, $V_c$ records its stored values or value format, and $W_c$ contains common words associated with it. Each analytical phrase $p\in\mathcal{P}$ is paired with a corresponding SQL pattern $\sigma_p$. For example, the severity entry includes
\begin{equation*}
\begin{aligned}
D_{\code{INJSVR}}
&= \text{``Most severe injury to any person involved in the crash''},\\
V_{\code{INJSVR}}
&= \{K,A,B,C,O\},\\
W_{\code{INJSVR}}
&= \{\textit{severity},\textit{severe},\textit{worst},
     \textit{injury},\textit{fatal},\dots\}.
\end{aligned}
\end{equation*}

Table~\ref{tab:kb} lists further examples of both parts. The entries combine three sources. Dictionary descriptions cover all $169$ columns. Stored codes and formats describe how those fields appear in the database; for low-cardinality categorical fields, read-only queries retrieve the observed values directly. This supplies exact representations such as \code{\textquotesingle{}MILWAUKEE\textquotesingle{}} rather than an alternative spelling such as \emph{Milwaukee County}. Manually curated word sets cover the $28$ frequently queried columns, allowing terms such as \emph{drunk} and \emph{OWI} to retrieve \code{ALCFLAG}. The convention entries add operational definitions that field descriptions alone do not provide. They specify, for example, how to express a factor that is not marked present, how to distinguish fatal crashes from fatality counts, and how to calculate a crash share. For this study, severity-based group rankings use the following documented convention:
\begin{equation}
\mathrm{Sev}(g)=\sum_{i\in g}w(s_i),
\qquad
w(K)=5,\quad w(A)=4,\quad w(B)=3,\quad
w(C)=2,\quad w(O)=1,
\label{eq:sev}
\end{equation}
where $s_i$ is the \code{KABCO} severity code of crash $i$ in group $g$. This score combines crash frequency and severity; it is an explicit analysis convention used by the workflow rather than a meaning inferred from an abbreviated column name.

\begin{table}[pos=t]
\caption{Examples of column knowledge and analytical conventions in $\K$.}
\label{tab:kb}
\centering\footnotesize
\renewcommand{\code}[1]{\texttt{#1}}
\setlength{\tabcolsep}{6pt}
\begin{tabular}{@{}llll@{}}
\toprule
\tabgroup{4}{(a) Column entries $\K_{\mathrm{col}}$}
\textbf{Column $c$} & \textbf{Description $D_c$} & \textbf{Stored values $V_c$} & \textbf{Associated words $W_c$} \\
\midrule
\code{CNTYNAME} & County of the crash
& $72$ names, e.g.\ \code{\textquotesingle{}MILWAUKEE\textquotesingle{}}
& \emph{county}, \emph{counties} \\
\code{ALCFLAG} & Police suspected alcohol use
& \code{\textquotesingle{}Y\textquotesingle{}}, \code{\textquotesingle{}UNKN\textquotesingle{}}, or \code{NULL}
& \emph{alcohol}, \emph{drunk}, \emph{OWI} \\
\code{CYCLFLAG} & Motorcycle involvement
& \code{\textquotesingle{}Y\textquotesingle{}} or \code{NULL}
& \emph{motorcycle}, \emph{motorbike} \\
\code{TOTFATL} & Number of fatalities
& Integer
& \emph{fatalities}, \emph{deaths}, \emph{killed} \\
\midrule
\tabgroup{4}{(b) Conventions $\K_{\mathrm{conv}}$}
\textbf{Phrase $p$} & \textbf{Operational interpretation}
& \multicolumn{2}{l}{\textbf{SQL pattern $\sigma_p$}} \\
\midrule
\emph{without} a factor & Factor not marked present
& \multicolumn{2}{l}{\code{\textit{flag} IS NULL OR \textit{flag} <> \textquotesingle{}Y\textquotesingle{}}} \\
\emph{fatal} & Crash involving at least one death
& \multicolumn{2}{l}{\code{TOTFATL > 0}} \\
\emph{share} of crashes & Percentage meeting a condition
& \multicolumn{2}{l}{\code{100 * AVG(CASE WHEN \textit{condition} THEN 1 ELSE 0 END)}} \\
\emph{worst} & Severity score defined in Eq.~\ref{eq:sev}
& \multicolumn{2}{l}{\code{SUM(CASE INJSVR WHEN \textquotesingle{}K\textquotesingle{} THEN 5 \dots\ END)}} \\
\bottomrule
\end{tabular}
\end{table}

Knowledge is retrieved according to each component's task. The Schema Linker uses descriptions and associated words to identify relevant columns. The Generator receives the selected column entries together with the analysis conventions, limiting the schema context included in its prompt. A deterministic normalization step subsequently aligns county literals with sufficiently similar stored spellings and leaves unmatched names unchanged. In the grounded CrashDB-QA+ design, the Judge and evidence-carrying Refiner receive entries for both the columns used in a query and related alternatives. For example, a motorcycle query that incorrectly uses the bicycle flag \code{BIKEFLAG} can be assessed alongside the definition of \code{CYCLFLAG}. The experiments separately examine the effects of grounding generation and verification.

\subsection{Question understanding}\label{sec:core}

Question understanding converts the analyst's request into an explicit analytical specification and retrieves the schema elements needed to express it. Keeping these outputs separate makes interpretation and field selection inspectable before SQL generation. Table~\ref{tab:prompts} summarizes the roles, inputs, and outputs of the four language-model agents used in Steps~1, 3, 6, and~8.

\begin{table}[pos=t]
\caption{Roles and prompts of the four language-model agents. Role instructions are quoted in abridged form.}
\label{tab:prompts}
\begin{tcbraster}[raster columns=2,raster equal height=rows,raster column skip=5pt,raster row skip=4pt]
\begin{tcolorbox}[agentcard=promptline,title={Step 1\enspace Interpreter}]
\cardrow{tagrole}{Role}{\emph{``You are the intent-parsing stage of a traffic-crash data assistant. Return a compact JSON object describing what the user wants. Do not write SQL.''}}
\tcblower
\cardrow{taginput}{Input}{The question and the intent schema.}
\vspace{1.5pt}\cardrow{tagoutput}{Output}{JSON intent $z$ with metric, filters, group, time, and rank.}
\end{tcolorbox}
\begin{tcolorbox}[agentcard=promptline,title={Step 3\enspace Generator}]
\cardrow{tagrole}{Role}{\emph{``You are the SQL-generation stage. Produce one read-only Oracle \code{SELECT} that answers the user's request. Output SQL only.''}}
\tcblower
\cardrow{taginput}{Input}{Table name; entries of the linked columns from $\K$; conventions and stored value formats; Oracle dialect rules; ten worked examples; the question, marked authoritative; the intent, marked as a hint; feedback $\varphi$ on a retry.}
\vspace{1.5pt}\cardrow{tagoutput}{Output}{One SQL query.}
\end{tcolorbox}
\begin{tcolorbox}[agentcard=plusline,title={Step 6\enspace Judge\hfill CrashDB-QA+}]
\cardrow{tagrole}{Role}{\emph{``You are the quality-control judge. Be strict: a query that runs but answers the wrong question must fail.''}}
\tcblower
\cardrow{taginput}{Input}{The question, intent, query, and result preview; deterministic warnings; entries in $\K$ of used and related unused columns; checks, and guardrails against invented requirements.}
\vspace{1.5pt}\cardrow{tagoutput}{Output}{JSON verdict, \textsc{pass} or \textsc{fail}, with four rubric scores, issues, and a suggested fix.}
\end{tcolorbox}
\begin{tcolorbox}[agentcard=promptline,title={Step 8\enspace Explainer}]
\cardrow{tagrole}{Role}{\emph{``You are explaining a query result to a transportation analyst who did not write SQL. Cite the key numbers and add one caveat. Do not invent numbers.''}}
\tcblower
\cardrow{taginput}{Input}{The question and a preview of the result.}
\vspace{1.5pt}\cardrow{tagoutput}{Output}{Two to four plain-language sentences.}
\end{tcolorbox}
\end{tcbraster}
\end{table}

\textbf{Step~1, the Interpreter.}
The Interpreter extracts five elements from the question:
\begin{equation}
\mathrm{Int}:\ q\ \longmapsto\
z=
\big(
\text{metric},\,
\text{filters},\,
\text{group},\,
\text{time},\,
\text{rank}
\big).
\label{eq:interp}
\end{equation}
These fields specify what to measure, which records to include, how to group them, the relevant period, and any requested ordering or limit. The output is a compact JSON object containing no SQL. For the severity-ranking example, it identifies a severity-weighted metric, county-level grouping, the specified year, and a top-five ranking. The structured intent supports schema retrieval and subsequent assessment, while the original question remains authoritative throughout the workflow.

\textbf{Step~2, the Schema Linker.}
The Schema Linker retrieves columns using lexical matches against their documented descriptions, associated words, and names. Let $T$ contain the terms extracted from the question and intent, and let $\tau(\cdot)$ tokenize text. Each column receives the score
\begin{equation}
\mathrm{score}(c)=
\left|T\cap\big(\tau(D_c)\cup W_c\big)\right|
+
\lambda\left|T\cap\tau(\mathrm{name}(c))\right|.
\label{eq:link}
\end{equation}
The first term measures semantic cues available in the knowledge base, while the second weights direct column-name matches by $\lambda$. The top $m$ columns form the linked set $\Sch_L$. A supplementary rule includes \code{INJSVR} whenever severity-related language is detected, ensuring that the severity score in Eq.~\ref{eq:sev} has its required input. Thus, the running example retrieves the county and severity fields together with the temporal fields needed to restrict the analysis period. Retrieval uses deterministic scoring and rules.

\subsection{Query generation and execution}\label{sec:construct}

The construction phase translates the question into SQL and establishes whether the query may access the database. Generation uses the retrieved domain knowledge; validation and execution apply deterministic controls independently of the model's instructions.

\textbf{Step~3, the Generator.}
The Generator receives the question, structured intent, linked column entries, analysis conventions, and any feedback from a previous attempt:
\begin{equation}
\mathrm{Gen}:
\big(q,z,\Sch_L,\K,\varphi\big)
\ \longmapsto\
\{Q_1,\dots,Q_k\}.
\label{eq:gen}
\end{equation}
Here, $\K$ denotes the relevant knowledge supplied to the prompt, and $\varphi$ is empty on the initial attempt. Prompt blocks follow a fixed order: table information, linked column entries, conventions and value formats, Oracle SQL rules, ten worked examples, the question, the intent, and retry feedback. The question is explicitly designated as authoritative, with the intent serving as supporting guidance. The first candidate uses greedy decoding; additional candidates, when enabled, are sampled and deduplicated. County literals are then normalized using the stored values described in Section~\ref{sec:rag}.

In a multi-candidate configuration, candidates that pass validation and execute successfully are grouped by their normalized results. The selected query belongs to the largest agreement group \citep{wang2023selfconsistency}:
\begin{equation}
i^\ast
=
\arg\max_i
\sum_j
\ind\!\big[\rows(R_i)=\rows(R_j)\big],
\qquad
\hat{Q}=Q_{i^\ast}.
\label{eq:vote}
\end{equation}
Result-based voting allows equivalent SQL formulations to support the same answer. Agreement is a selection criterion rather than a correctness guarantee, because multiple candidates can reproduce the same analytical error. With one candidate, the workflow uses that candidate if it passes validation and execution.

\textbf{Step~4, the Validator.}
Every candidate is checked before database execution:
\begin{equation}
\begin{aligned}
\mathrm{Val}(Q)
={}&
\ind\!\big[Q\text{ is a single read-only statement}\big]\\
&\wedge
\ind\!\big[\mathrm{cols}(Q)\subseteq\Sch\big]\\
&\wedge
\ind\!\big[Q\text{ parses in the Oracle dialect}\big].
\end{aligned}
\label{eq:val}
\end{equation}
The first check admits a single statement beginning with \code{SELECT} or \code{WITH} and rejects prohibited write operations. The second checks referenced identifiers against the schema while accounting for SQL syntax and quoted values. The third checks dialect compatibility. These controls reject malformed or inadmissible queries; they do not determine whether the query expresses the intended analysis. If no candidate passes, validation errors become feedback $\varphi$ for another generation attempt, subject to the configured attempt limit $N$.

\textbf{Step~5, the Executor.}
The Executor runs each admitted query within the agency's database environment:
\begin{equation}
R=\exec(\hat{Q},\Db).
\label{eq:exec}
\end{equation}
It repeats the read-only check immediately before database access and applies limits on execution time and returned rows. Execution errors are recorded and returned as feedback for revision. Connection and query times are logged separately to support latency analysis. Model inference and result processing remain local, so crash records are not transmitted to an external model service. In the core workflow, the selected result proceeds directly to Step~8; CrashDB-QA+ first applies the verification and refinement stages below.

\subsection{Result verification and explanation}\label{sec:verify}

CrashDB-QA+ extends admissibility checks with an assessment of analytical alignment. Its Judge examines the question, generated query, and execution evidence, while its Refiner converts identified problems into feedback for another attempt. This loop estimates correctness without access to the reference answers used in evaluation.

\textbf{Step~6, the Judge.}
The Judge returns a verdict $v$ and a set of identified issues $\chi$:
\begin{equation}
\mathrm{Judge}:
\big(q,z,\hat{Q},R;\K_J\big)
\ \longmapsto\
(v,\chi),
\label{eq:judgemap}
\end{equation}
where $\K_J$ denotes the retrieved schema knowledge supplied in the grounded design and is omitted in the ungrounded design. The LLM rubric evaluates four dimensions: relevance, aggregation, filters, and completeness. It receives the question, intent, SQL, and a preview of the returned rows, and provides a suggested correction when it detects a mismatch. The question remains the basis for assessment: the Judge is instructed to avoid introducing requirements that the analyst did not request. Grounding adds definitions for used columns and related alternatives, helping the Judge distinguish fields whose abbreviated names may otherwise be confused.

Optional deterministic checks flag potentially suspicious result structures: an empty table, an entirely \code{NULL} column, an all-zero numeric column, a single-row breakdown or trend, or a seasonal breakdown containing more than four groups. When enabled, these checks act as vetoes:
\begin{equation}
v=
\ind\!\big[\text{LLM rubric}=\textsc{pass}\big]
\wedge
\ind\!\big[\text{no enabled deterministic check fires}\big].
\label{eq:judge}
\end{equation}
These rules are heuristics rather than proofs of error, because valid analyses can also return empty, zero-valued, or single-row results. Their effects are therefore evaluated separately from those of the LLM Judge. In the running example, a raw-count ranking can pass the structural checks, while the rubric can reject it for omitting the requested severity weighting.

\textbf{Step~7, the Refiner.}
When verification fails, the Refiner assembles feedback for the Generator:
\begin{equation}
\mathrm{Ref}:
\big(q,\hat{Q},R,\chi;\K\big)
\ \longmapsto\
\varphi.
\label{eq:ref}
\end{equation}
The Refiner is deterministic and makes no additional model call. We compare two feedback designs:
\begin{equation}
\begin{aligned}
\varphi_{\mathrm{crit}}
&=
\big(\hat{Q},\chi\big),\\
\varphi_{\mathrm{evid}}
&=
\big(
\hat{Q},\chi,
\text{result evidence},
\text{column facts},
\text{repair pattern},
\text{tried queries}
\big).
\end{aligned}
\label{eq:refine}
\end{equation}
The \emph{critique-only} design forwards the failed query and the Judge's issues and suggested correction. The \emph{evidence-carrying} design additionally provides the result shape, initial rows, and warnings about repeated output values; relevant column definitions; a targeted SQL repair pattern; and previously attempted queries. Repair patterns address three recurring structures: seasonal bucketing, two-group comparisons, and breakdowns that collapse into an aggregate. The attempted-query history is included to discourage repetition. These additions allow the Generator to revise a query using both the critique and the evidence that motivated it.

Feedback returns the workflow to Step~3. Generation, validation, execution, and verification continue until a result passes or the iteration limit $N$ is reached. If no result passes, the critique-only configuration selects the final attempt, while the evidence-carrying configuration selects the attempt with the fewest unresolved issues. The trace records that verification was not passed. The iteration limit bounds computation and latency; the experiments assess whether the resulting revisions repair errors or degrade previously correct answers.

\textbf{Step~8, the Explainer.}
The Explainer converts the selected result into a concise answer:
\begin{equation}
\mathrm{Expl}:
\big(q,R\big)
\ \longmapsto\
a.
\label{eq:expl}
\end{equation}
It receives the question and a preview of the result and produces two to four sentences stating the answer, reporting key values, and identifying a relevant limitation. Its instructions restrict numerical claims to values supplied in the result. Explanation is the final presentation stage: it communicates the selected output but does not independently establish query correctness.

\begin{figure}[pos=tbp]
\centering
\includegraphics[width=0.9\linewidth]{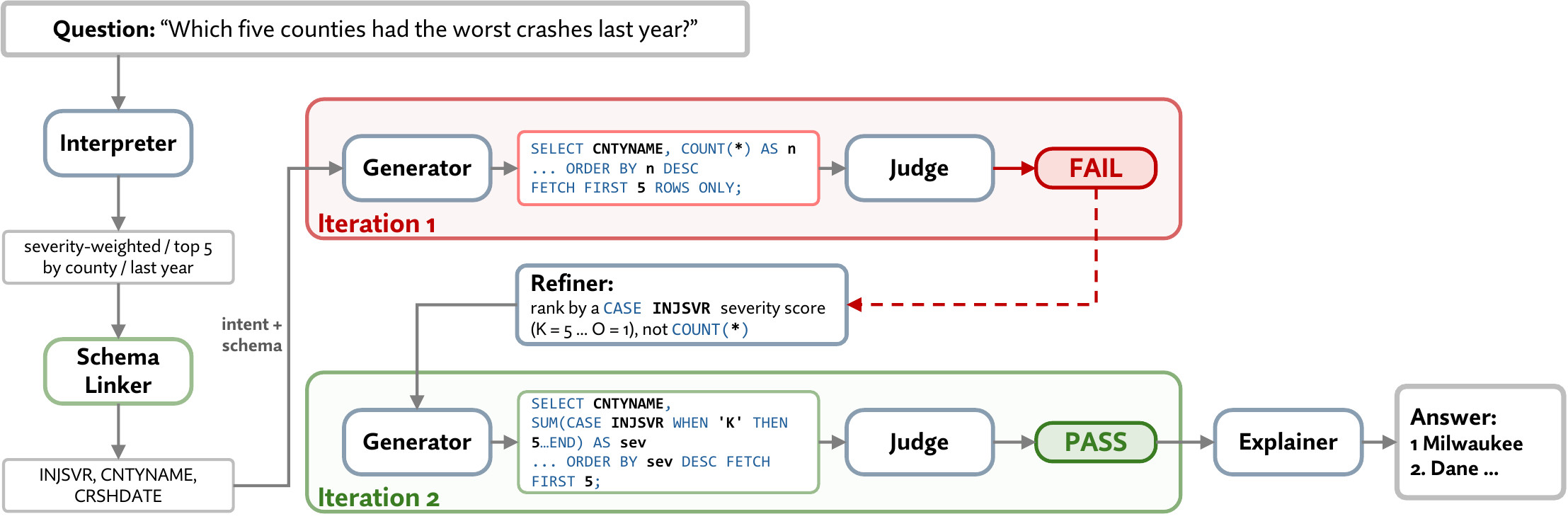}
\caption{Example of execution-informed refinement. An admissible query initially ranks counties by crash count. The Judge identifies the mismatch with the requested severity-weighted measure, and refinement produces a revised query that passes verification on the second iteration. The dashed red path marks the refinement loop.}
\label{fig:trace}
\end{figure}

Figure~\ref{fig:trace} illustrates the complete workflow using the severity-ranking example. The initial query ranks counties by crash count and passes validation because it is admissible SQL. The Judge identifies the missing severity weighting, and the Refiner supplies feedback for revision. The Generator then constructs the weighted aggregation in Eq.~\ref{eq:sev}, after which the revised result passes verification and proceeds to explanation. The trace demonstrates the distinct roles of validation, semantic assessment, and evidence-informed repair.

\section{The CrashDB-QA Benchmark}\label{sec:benchmark}

The CrashDB-QA benchmark evaluates whether a natural-language analysis system can translate an analyst's request into the intended result under a documented crash-data schema. Cross-domain benchmarks such as Spider and BIRD assess performance across diverse databases \citep{yu2018spider,li2023bird}. Our benchmark complements this perspective by examining analytical depth within one governed database: whether a system correctly interprets coded fields, applies domain conventions, and combines operations required for safety analysis. Each question is paired with an analyst-reviewed reference query and its execution result on a frozen database snapshot. The benchmark therefore evaluates the complete path from a natural-language request to a computed answer, with particular attention to executable-but-wrong queries.

\subsection{Analytical taxonomy}

\begin{figure}[pos=tbp]
\centering
\includegraphics[width=0.9\linewidth]{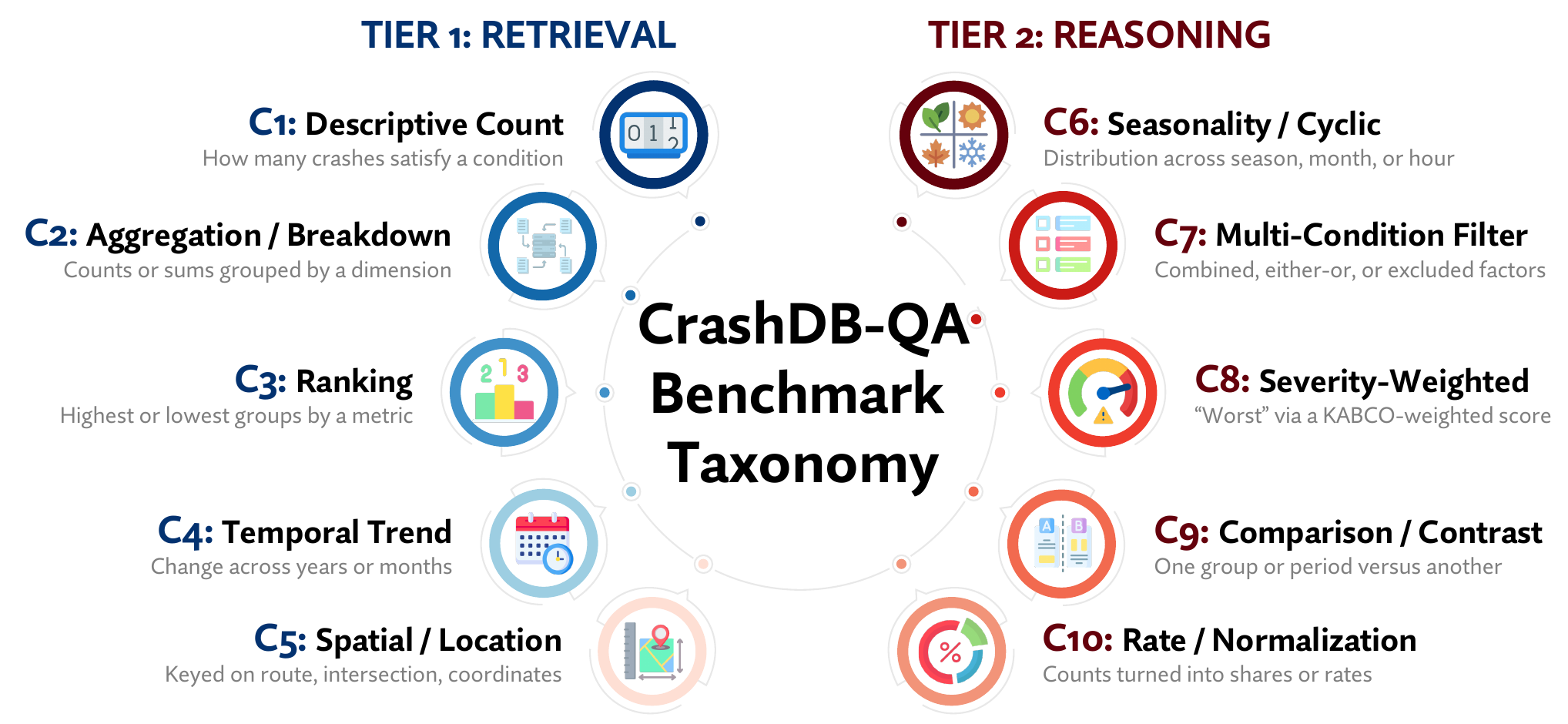}
\caption{Benchmark taxonomy and representative executable-but-wrong errors. Categories C1--C5 emphasize standard retrieval and aggregation, while C6--C10 require additional composition of conditions, groupings, or derived quantities.}
\label{fig:taxonomy}
\end{figure}

The benchmark organizes questions into ten categories according to their primary analytical operation, as summarized in Table~\ref{tab:qexamples} and Figure~\ref{fig:taxonomy}. Categories C1--C5 form the \emph{retrieval tier}, covering counts, grouped summaries, rankings, temporal trends, and location-based summaries. These tasks primarily apply standard filtering and aggregation to existing fields. Categories C6--C10 form the \emph{reasoning tier}, covering seasonal grouping, interacting conditions, severity-weighted measures, explicit comparisons, and normalized quantities. These tasks place additional demands on the composition of predicates, transformations, or derived measures. The tier distinction describes the analytical structure of the questions; retrieval tasks can still require multiple SQL operations, and both tiers depend on correct domain interpretation. This organization follows the broader distinction between recognition and reasoning used in the CrashSight video benchmark \citep{gan2026crashsight}, adapted here to structured database analysis.

\begin{table}[pos=t]
\caption{Representative questions across the ten benchmark categories.}
\label{tab:qexamples}
\centering\small
\setlength{\tabcolsep}{6pt}
\begin{tabular*}{\textwidth}{@{\extracolsep{\fill}}llp{11.2cm}@{}}
\toprule
\textbf{Code} & \textbf{Category} & \textbf{Representative question} \\
\midrule
\tabgroup{3}{Tier 1: retrieval through standard filtering and aggregation}
C1  & Descriptive count
& How many alcohol-related crashes occurred in Dane County in 2023? \\
C2  & Aggregation / breakdown
& How many crashes occurred in each county in 2023? \\
C3  & Ranking / top-$N$
& Which five counties had the most crashes in 2023? \\
C4  & Temporal trend
& How has the number of fatal crashes changed year over year? \\
C5  & Spatial / location
& How many crashes occurred at each intersection type in Dane County? \\
\midrule
\tabgroup{3}{Tier 2: reasoning through additional analytical composition}
C6  & Seasonality / cyclic
& How are pedestrian crashes distributed across the four seasons? \\
C7  & Multi-condition filter
& How many crashes involved speeding but not alcohol in a work zone? \\
C8  & Severity-weighted
& Which ten counties had the worst crashes by injury severity in 2023? \\
C9  & Comparison / contrast
& Did fatal crashes in Dane County rise or fall from 2022 to 2023? \\
C10 & Rate / normalization
& What share of crashes in each county involved a distracted driver? \\
\bottomrule
\end{tabular*}
\end{table}

The categories support diagnosis beyond an overall accuracy score. A seasonal question, for example, requires months to be assigned to the intended seasonal groups; a comparison requires distinct conditions for the quantities being contrasted; and a crash share requires an appropriate numerator and denominator. Errors in these operations can leave the query executable while changing the meaning of its result. Category-level performance therefore helps identify recurring analytical weaknesses, although each score reflects the combined effects of interpretation, schema linking, and query construction rather than an isolated model capability. Figure~\ref{fig:taxonomy} associates each category with a representative failure pattern.

\subsection{Question construction and development separation}

Questions combine operational requests with expert-authored cases. The operational requests originate from Community Maps inquiries routed to TOPS and WisDOT specialists and were anonymized and reformulated as benchmark questions. A domain expert supplied additional questions to cover analytical categories that occurred less frequently in those requests. Most questions specify a county and year, reflecting common safety-analysis needs while keeping reference results manageable for manual review. The final benchmark contains $300$ questions, with $30$ in each category. This balanced design gives every category equal weight in overall accuracy and supports direct comparison across analytical tasks; it does not estimate the frequency of those tasks in routine agency use.

Initial prompt wording, knowledge-base rules, and synonym mappings were developed using a separate set of questions excluded from benchmark scoring. The $300$ benchmark questions and their reference results were then fixed for evaluation. Section~\ref{sec:experiments} distinguishes the initial model comparisons and pre-registered verification configurations from subsequent diagnostic improvements motivated by observed failures. This distinction preserves the separation between development and initial evaluation while making clear where later analyses reuse the benchmark to investigate and address specific weaknesses.

\subsection{Reference construction and quality control}

Reference queries were constructed through a draft--review--audit process. A language model produced an initial SQL draft for each question, which two analysts independently reviewed and corrected. When their corrected queries returned different results, a third reviewer adjudicated the discrepancy. All references were executed against the frozen database snapshot before finalization. A random $15\%$ were additionally checked against raw records, and a subset was compared with published WisDOT crash statistics. These checks assessed both the intended analytical operation and its consistency with the underlying data.

Execution testing was particularly important for detecting errors in stored-value assumptions. Nine draft references tested contributing-factor flags against \code{'N'}, although that value was not present in the relevant fields. These queries were corrected using the documented coding conventions described in Section~\ref{sec:rag}. The resulting references therefore combine expert interpretation with checks against actual database representations. Using the same frozen snapshot for every reference and candidate query also prevents changes in the underlying records from affecting comparisons.

\subsection{Execution-based grading}

Candidate answers are graded using execution match, as defined in Eq.~\ref{eq:em}. A candidate is correct when its normalized result matches the reference result, regardless of differences in SQL formulation. This criterion accommodates equivalent implementations and directly tests the values returned to the analyst. Ranking references include the ranking measure alongside the selected locations or groups, allowing grading to compare both the selected entities and their associated values. Because the metric compares row multisets, it does not independently assess presentation order. The score also evaluates the computed result rather than the factual accuracy or clarity of the Explainer's prose.

Two considerations bound this evaluation. First, different interpretations of a question can produce different defensible answers. The benchmark fixes the intended interpretation through documented conventions: for example, \emph{worst} refers to the severity-weighted measure in Eq.~\ref{eq:sev}, and \emph{last year} refers to the last full calendar year. Scores therefore measure agreement with these operational definitions. Second, an incorrect query can coincidentally return the reference result on a single snapshot. Expert review and reference audits strengthen the ground truth, but cannot rule out such coincidences in candidate queries. Evaluation across multiple database instances or targeted test suites would provide stronger evidence of semantic equivalence \citep{zhong2020testsuite}. The benchmark consequently measures execution agreement on a fixed real-world dataset, with category-level analysis used to explain the observed successes and failures.

\section{Results and Discussion}\label{sec:experiments}

The evaluation examines whether CrashDB-QA can provide accurate answers within practical local-computing constraints, where its remaining errors arise, and whether execution-informed verification can repair them. We first compare the grounded core across nine open-weight models, then analyze latency and failure patterns. Case studies illustrate how valid SQL can produce an incorrect analytical answer. Finally, controlled verification experiments distinguish the Judge's ability to detect errors from the Refiner's ability to correct them. The model comparison uses one greedy run per model; the verification study uses the strongest model identified in that comparison.

\subsection{Experimental setup}\label{sec:eval-setup}

All experiments run locally through Ollama against a frozen DT4000 snapshot containing $4.4$ million crash records. Nine open-weight models span approximately 1B--8B parameters and include general-purpose, code-specialized, and reasoning-oriented variants. The general-purpose models are Gemma-3 1B and 4B \citep{gemma2025gemma3}, Llama-3.2 3B and Llama-3.1 8B \citep{grattafiori2024llama3}, and Qwen-2.5 7B \citep{qwen2024qwen25}. The code-specialized models are DeepSeek-Coder 6.7B \citep{guo2024deepseekcoder} and Qwen-2.5-Coder 7B \citep{hui2024qwen25coder}. The reasoning-oriented models are Qwen-3.5 4B and DeepSeek-R1 8B \citep{guo2025deepseekr1}. Each model is evaluated on all $300$ benchmark questions using the same prompt templates, grounding resources, reference answers, and database snapshot. Accuracy is measured by execution match under Eq.~\ref{eq:em}. Full traces record intermediate outputs, agent inference times, and database connection and execution times. Repeated runs of two models reproduced their category-level scores exactly under the same configuration; broader variation across seeds and runtime settings is outside the present evaluation.

The Schema Linker uses $\lambda=0.25$ and retrieves the top $m=16$ columns, supplemented by the rules described in Section~\ref{sec:core}. The Executor limits query time to $30$~s and returned results to $5{,}000$ rows. The Judge receives the first $10$ result rows, and the Explainer receives the first $20$. Single-candidate generation uses greedy decoding; additional candidates, when enabled, are sampled at temperature $0.4$. Verification runs use at most $N=3$ iterations. The initial model comparison evaluates the grounded core without semantic verification. Subsequent experiments first test the pre-registered verification configurations and then examine diagnostic improvements motivated by their observed failures.

\subsection{Accuracy across models}\label{sec:eval-sweep}

\begin{figure}[pos=t]
\centering
\includegraphics[width=0.85\linewidth]{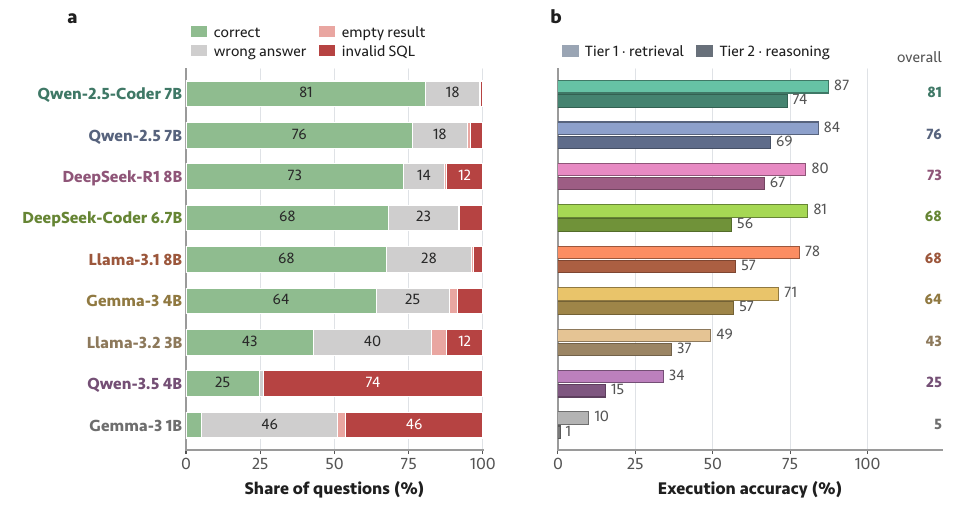}
\caption{Performance of nine local models on the $300$-question benchmark. (a) Question outcomes. (b) Accuracy on the retrieval and reasoning tiers, with overall accuracy shown at right.}
\label{fig:models}
\end{figure}

\begin{figure}[pos=t]
\centering
\includegraphics[width=0.9\linewidth]{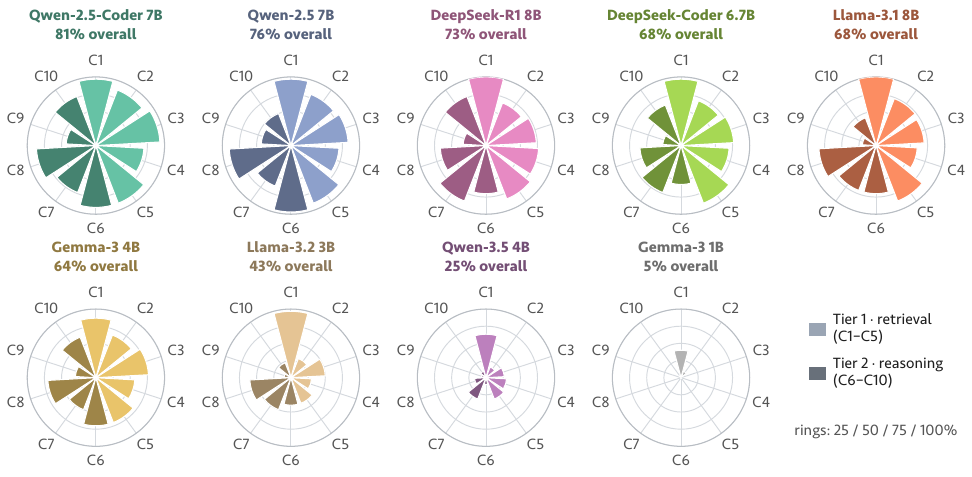}
\caption{Category-level accuracy by model, ordered by overall performance. Each circle shows retrieval categories on the right and reasoning categories on the left; concentric rings indicate $25$-percentage-point increments.}
\label{fig:cats}
\end{figure}

Table~\ref{tab:results} reports overall and category-level execution accuracy, while Figures~\ref{fig:models} and~\ref{fig:cats} summarize model and tier differences. Qwen-2.5-Coder 7B achieves the highest overall accuracy, $80.7\%$, followed by Qwen-2.5 7B at $76.3\%$ and DeepSeek-R1 8B at $73.3\%$. Performance varies substantially within similar parameter ranges, indicating that model size alone does not determine suitability for the workflow.

\begin{table}[pos=t]
\caption{Execution accuracy of the grounded CrashDB-QA core by model and analytical category (\%). Each category contains $30$ questions, for $300$ questions in total. Inv.\ denotes the number of questions producing non-executable queries.}
\label{tab:results}
\centering\small
\setlength{\tabcolsep}{2pt}
\begin{tabular}{@{}llrrrrrrrrrrrr@{}}
\toprule
{} & {} & \multicolumn{5}{c}{\textbf{Tier 1: Retrieval}} & \multicolumn{5}{c}{\textbf{Tier 2: Reasoning}} & {} & {} \\
\cmidrule(lr){3-7}\cmidrule(lr){8-12}
\textbf{Model} & \textbf{Regime} & \textbf{C1} & \textbf{C2} & \textbf{C3} & \textbf{C4} & \textbf{C5} & \textbf{C6} & \textbf{C7} & \textbf{C8} & \textbf{C9} & \textbf{C10} & \textbf{Overall} & \textbf{Inv.} \\
\midrule
Qwen-2.5-Coder 7B   & code      & \heat{96.7} & \heat{86.7} & \heat{93.3} & \heat{70.0} & \heat{90.0} & \heat{90.0} & \heat{73.3} & \heat{86.7} & \heat{43.3} & \heat{76.7} & \heat{80.7} & \textbf{3} \\
Qwen-2.5 7B         & general   & \heat{96.7} & \heat{80.0} & \heat{83.3} & \heat{70.0} & \heat{90.0} & \heat{96.7} & \heat{63.3} & \heat{90.0} & \heat{43.3} & \heat{50.0} & \heat{76.3} & 13 \\
DeepSeek-R1 8B      & reasoning & \heat{100.0} & \heat{66.7} & \heat{73.3} & \heat{76.7} & \heat{83.3} & \heat{70.0} & \heat{86.7} & \heat{66.7} & \heat{33.3} & \heat{76.7} & \heat{73.3} & 37 \\
DeepSeek-Coder 6.7B & code      & \heat{96.7} & \heat{70.0} & \heat{76.7} & \heat{70.0} & \heat{90.0} & \heat{56.7} & \heat{73.3} & \heat{60.0} & \heat{26.7} & \heat{63.3} & \heat{68.3} & 24 \\
Llama-3.1 8B        & general   & \heat{100.0} & \heat{73.3} & \heat{70.0} & \heat{60.0} & \heat{86.7} & \heat{70.0} & \heat{70.0} & \heat{83.3} & \heat{20.0} & \heat{43.3} & \heat{67.7} & 10 \\
Gemma-3 4B          & general   & \heat{86.7} & \heat{66.7} & \heat{76.7} & \heat{56.7} & \heat{70.0} & \heat{70.0} & \heat{50.0} & \heat{70.0} & \heat{30.0} & \heat{63.3} & \heat{64.0} & 26 \\
Llama-3.2 3B        & general   & \heat{96.7} & \heat{30.0} & \heat{50.0} & \heat{30.0} & \heat{40.0} & \heat{40.0} & \heat{50.0} & \heat{60.0} & \heat{10.0} & \heat{23.3} & \heat{43.0} & 37 \\
\addlinespace[2pt]
Qwen-3.5 4B         & reasoning & \heat{63.3} & \heat{16.7} & \heat{26.7} & \heat{30.0} & \heat{33.3} & \heat{10.0} & \heat{33.3} & \heat{16.7} & \heat{6.7} & \heat{10.0} & \heat{24.7} & 222 \\
Gemma-3 1B          & general   & \heat{40.0} & \heat{10.0} & \heat{0.0} & \heat{0.0} & \heat{0.0} & \heat{0.0} & \heat{0.0} & \heat{0.0} & \heat{0.0} & \heat{3.3} & \heat{5.3} & 139 \\
\bottomrule
\end{tabular}
\end{table}

\noindent\textbf{Model size and executable output.}
The results indicate a substantial capability gap between the smallest model and the stronger local models. Gemma-3 improves from $5.3\%$ at 1B parameters to $64.0\%$ at 4B, while the evaluated models in the approximately 7B--8B range achieve $67.7$--$80.7\%$, as Figure~\ref{fig:size}a shows. However, Qwen-3.5 4B reaches only $24.7\%$, illustrating that size does not ensure compatibility with the task and output contract. Non-executable output is a major contributor to poor performance: Gemma-3 1B produces $139$ non-executable queries, and Qwen-3.5 4B produces $222$. Grounding supplies domain information, but cannot compensate for a model that frequently fails to produce usable SQL under the evaluated configuration.

\noindent\textbf{Retrieval and analytical composition.}
Across the seven models scoring at least $43\%$ overall, reasoning-tier accuracy is approximately $13$--$25$ percentage points below retrieval-tier accuracy. Qwen-2.5-Coder 7B, for example, achieves $87.3\%$ on retrieval and $74.0\%$ on reasoning. This pattern is consistent with the broader recognition--reasoning distinction observed in crash-video understanding \citep{gan2026crashsight}, although the tasks and evaluation settings differ. The database results show that identifying relevant records and applying standard aggregations are more reliable than composing additional analytical operations. The gap also varies by category: seasonality and severity weighting can be handled accurately by the stronger models, while explicit comparisons remain difficult.

\noindent\textbf{Comparisons as a persistent weakness.}
No model exceeds $43.3\%$ on C9. The strongest code-specialized model and its general-purpose counterpart both reach this value, despite their different overall scores. The $30$ comparison questions include two-period pivots, population contrasts, comparisons of shares, and severity-score contrasts. These tasks require separate quantities to be computed under distinct conditions and then combined into the requested comparison. Their consistently low scores identify this form of composition as a shared weakness across the evaluated configurations. As shown in Section~\ref{sec:eval-loop}, the verification loop also fails to repair these questions in the present study.

\noindent\textbf{Specialization and output discipline.}
The same-family comparison between Qwen-2.5-Coder 7B and Qwen-2.5 7B provides the clearest evidence of an advantage associated with code specialization. The code model improves overall accuracy by $4.4$ percentage points and reduces non-executable queries from $13$ to $3$. Its largest category-level gain is in rates and normalization, although it does not improve every category. Across families, DeepSeek-Coder 6.7B achieves performance similar to Llama-3.1 8B, at $68.3\%$ and $67.7\%$, respectively. The reasoning-oriented models do not improve the overall accuracy frontier: DeepSeek-R1 8B reaches $73.3\%$, and both reasoning-oriented models exhibit output-format failures involving deliberative text. These comparisons support the practical value of code-oriented models in this workflow, while differences in training data and model design prevent attributing all performance differences to training regime alone.

\subsection{Latency and cost}\label{sec:eval-cost}

\begin{figure}[pos=t]
\centering
\includegraphics[width=0.8\linewidth]{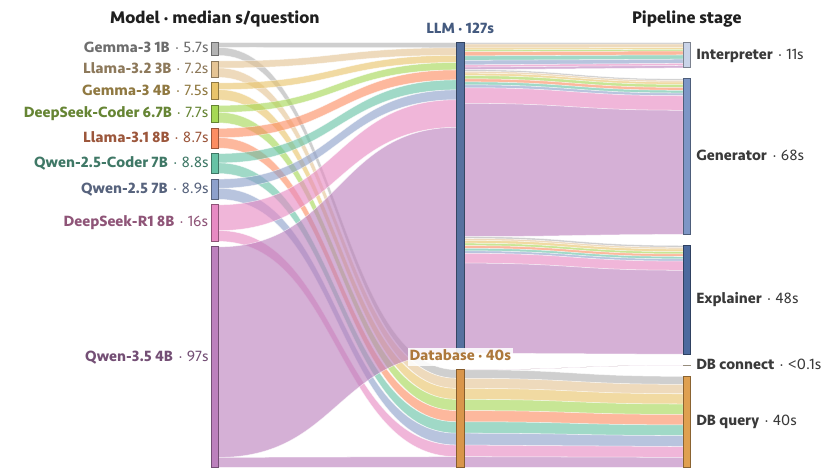}
\caption{Stage-level latency across local models. Ribbon widths represent time attributed to pipeline stages before and after database execution.}
\label{fig:cost}
\end{figure}

\begin{figure}[pos=t]
\centering
\begin{minipage}[t]{0.48\linewidth}
\centering
\includegraphics[width=\linewidth]{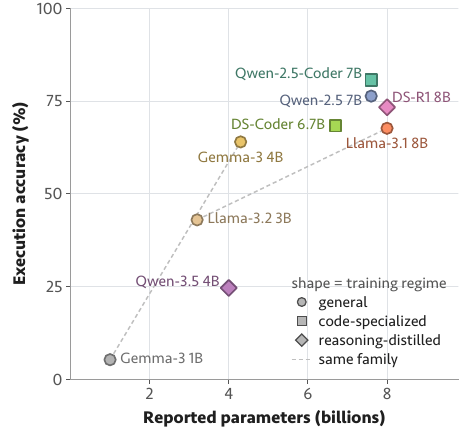}\\[-1pt]
{\footnotesize (a)}
\end{minipage}\hfill
\begin{minipage}[t]{0.48\linewidth}
\centering
\includegraphics[width=\linewidth]{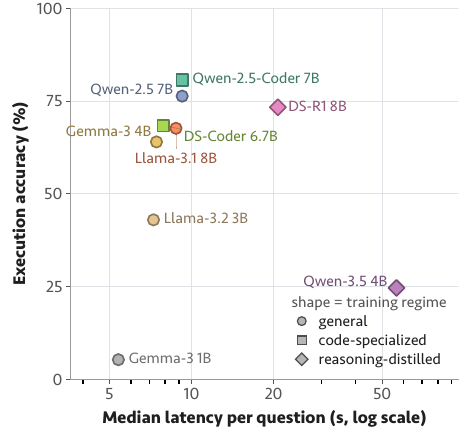}\\[-1pt]
{\footnotesize (b)}
\end{minipage}
\caption{Execution accuracy versus (a) parameter count and (b) median end-to-end latency on a logarithmic scale. Colors identify models, and marker shapes identify training regimes.}
\label{fig:size}
\label{fig:pareto}
\end{figure}

Accuracy determines whether an answer is useful, while latency determines whether the workflow supports interactive analysis. Figure~\ref{fig:cost} decomposes time by pipeline stage, and Figure~\ref{fig:pareto} relates end-to-end latency to accuracy. Database execution time varies relatively little among models that reliably reach execution: median query time is $4.1$--$4.8$~s, with negligible connection overhead. Most differences in end-to-end latency therefore arise from model inference and workflow behavior. The two lowest-performing models reach execution less consistently, so their database-time summaries should be interpreted alongside their high failure rates.

\noindent\textbf{Inference time and latency variability.}
Median model time is approximately $3$--$4$~s for the stronger general-purpose and code-specialized models, compared with $11.3$~s and $56.2$~s for the two reasoning-oriented models. The latter also exhibit long latency tails, with 95th-percentile end-to-end times of approximately $475$~s and $110$~s, respectively, compared with roughly $36$--$40$~s for the other models. The additional inference time does not produce a corresponding improvement in execution accuracy under this configuration.

\noindent\textbf{Accuracy--latency tradeoff.}
Qwen-2.5-Coder 7B provides the highest accuracy at a median end-to-end latency of $9.2$~s. No evaluated model is both faster and more accurate. DeepSeek-Coder 6.7B is approximately $1.4$~s faster but scores $12.4$ percentage points lower, while the reasoning-oriented models require more time for lower accuracy. Thus, Qwen-2.5-Coder 7B provides a favorable operating point for the subsequent verification study. These timings characterize the evaluated local configuration; they are not hardware-independent performance estimates.

\subsection{Error analysis}\label{sec:eval-errors}

We inspected incorrect outputs to distinguish failures in analytical construction from failures in domain interpretation and executable output. Among the stronger models, compositional errors were the predominant remaining pattern. Queries often selected appropriate fields and records but combined them incorrectly: a missing \code{GROUP BY} collapsed a breakdown into a single value, an unnecessary row limit truncated a trend, incorrectly grouped \code{AND}/\code{OR} conditions changed the selected population, or a seasonal request was answered with monthly counts. Comparison queries also reused the same predicate for quantities that should have been computed separately. These failures explain why high schema familiarity does not translate directly into reliable analytical composition.

Two additional error classes concern domain grounding. \emph{Value-grounding errors} use an incorrect stored representation, such as an unmatched county name or a flag comparison against \code{'N'}. Such queries can execute and return an empty or zero-valued result, making the error difficult to distinguish from a legitimate absence of crashes. The knowledge base and county normalization address these failures. \emph{Column-selection errors} use a valid but semantically inappropriate field, such as a commercial-vehicle or bicycle flag in a motorcycle analysis. These errors motivate supplying the Judge with definitions of related alternative columns, as examined in Section~\ref{sec:eval-loop}.

Some requests also depend on information unavailable in the crash table, such as an external population denominator for a population-normalized rate. Such cases require recognizing that the available data do not support the requested calculation. This differs from computing a share within the crash records, where both numerator and denominator can be obtained from the table. Finally, \emph{non-executable output} remains a separate failure mode. It affects $139$, $37$, and $222$ questions for Gemma-3 1B, DeepSeek-R1 8B, and Qwen-3.5 4B, respectively, compared with only $3$ for Qwen-2.5-Coder 7B. For the latter model, most incorrect answers therefore arise after the requirement to produce executable SQL has already been met.

\subsection{Case studies}\label{sec:eval-cases}

Two outputs from Qwen-2.5-Coder 7B illustrate the distinction between correct domain-grounded composition and executable analytical error. The first correctly applies the benchmark's severity convention. The second retrieves the appropriate records but uses the wrong temporal grouping, after which the explanation introduces additional inaccuracies. The examples include the executed SQL, result excerpts, and recorded Explainer outputs.

These cases show why admissibility, analytical correctness, and explanation quality require separate attention. The seasonal query passes validation because its syntax, fields, and access mode are acceptable; detecting the error requires relating its grouping operation and returned rows to the question. The explanation can then compound the problem by presenting unsupported or incorrect conclusions. Verification should therefore inspect the query together with execution evidence, while recognizing that a correct result alone does not guarantee a faithful explanation. The following experiments assess the query-repair component of this problem.

\begin{qacorrect}{Correct: C8 severity-weighted county ranking}
\qalabel{Question.}
Which 10 counties had the worst crashes by injury severity in 2023?

\qalabel{Model query}
(identical to the reference query):
\begin{lstlisting}[style=crashsql]
SELECT CNTYNAME,
       SUM(CASE INJSVR WHEN 'K' THEN 5 WHEN 'A' THEN 4
                       WHEN 'B' THEN 3 WHEN 'C' THEN 2
                       ELSE 1 END) AS severity_score
FROM   CRASH_TABLE
WHERE  CRSHYEAR = 2023
GROUP BY CNTYNAME
ORDER BY severity_score DESC
FETCH FIRST 10 ROWS ONLY
\end{lstlisting}

\qalabel{Result excerpt.}
MILWAUKEE 33{,}577; DANE 12{,}116; WAUKESHA 9{,}996; \dots\ ($10$ rows).

\qalabel{Recorded Explainer output.}
\emph{``The top 10 counties with the worst crashes by injury severity
in 2023 [\dots] are: Milwaukee (33{,}577), Dane (12{,}116), Waukesha (9{,}996), \dots\ Caveat:
the data [\dots] may not include all types of injuries or other relevant factors affecting crash
severity.''}

\qalabel{Interpretation.}
The query applies the specified \code{KABCO} weights, aggregates by county, restricts the year, and returns the ten highest-scoring counties. Its execution result matches the reference. The explanation reports the leading values, although its caveat is generic rather than specific to the frequency--severity interpretation of the score.
\end{qacorrect}

\begin{qawrong}{Incorrect: C6 monthly counts in place of seasonal totals}
\qalabel{Question.}
How many pedestrian crashes occurred in each season in Milwaukee County in 2023?

\qalabel{Reference query}
(groups months into four seasons):
\begin{lstlisting}[style=crashsql]
SELECT CASE WHEN EXTRACT(MONTH FROM CRSHDATE) IN (3,4,5)  THEN 'spring'
            WHEN EXTRACT(MONTH FROM CRSHDATE) IN (6,7,8)  THEN 'summer'
            WHEN EXTRACT(MONTH FROM CRSHDATE) IN (9,10,11) THEN 'fall'
            ELSE 'winter' END AS season,
       COUNT(*) AS ped_crashes
FROM   CRASH_TABLE
WHERE  PEDFLAG = 'Y' AND CNTYNAME = 'MILWAUKEE' AND CRSHYEAR = 2023
GROUP BY CASE ... END      -- the same season expression
ORDER BY ped_crashes DESC
\end{lstlisting}

\qalabel{Reference result.}
Summer $141$; fall $141$; spring $130$; winter $123$.

\qalabel{Model query}
(groups by calendar month):
\begin{lstlisting}[style=crashsql]
SELECT EXTRACT(MONTH FROM CRSHDATE) AS crash_month,
       COUNT(*) AS crash_count
FROM   CRASH_TABLE
WHERE  CNTYNAME = 'MILWAUKEE' AND PEDFLAG = 'Y' AND CRSHYEAR = 2023
GROUP BY EXTRACT(MONTH FROM CRSHDATE)
\end{lstlisting}

\qalabel{Model result excerpt.}
$12$ monthly rows: January $32$, February $39$, March $40$, \dots.

\qalabel{Recorded Explainer output.}
\emph{``In Milwaukee County in 2023, there were a total of 544
pedestrian crashes. The highest number of crashes occurred in the summer months (July and
August), with 58 and 50 crashes respectively. The lowest number of crashes was in February, with
only 39 incidents.''}

\qalabel{Interpretation.}
The query selects the correct population but returns monthly counts rather than the four requested seasonal totals. The explanation does not resolve this mismatch and adds errors: the stated total of $544$ differs from the reference total of $535$, and February is described as the minimum despite January having fewer crashes.
\end{qawrong}

\subsection{Verification loop}\label{sec:eval-loop}

The verification study separates two requirements for useful refinement. \textbf{RQ1} asks whether the Judge distinguishes correct from incorrect results. \textbf{RQ2} asks whether the resulting feedback improves final execution accuracy without introducing excessive regressions. Both experiments use Qwen-2.5-Coder 7B. RQ1 replays alternative Judge designs over recorded core outputs with known execution-match labels. RQ2 runs the complete workflow on all $300$ questions, first with the pre-registered configurations and then with diagnostic improvements. Paired accuracy differences are assessed using exact McNemar tests. The analysis also uses $95\%$ percentile bootstrap intervals and applies Holm--Bonferroni correction separately within the pre-registered comparison family and the subsequent improvement family.

\begin{figure}[pos=t]
\centering
\includegraphics[width=\linewidth]{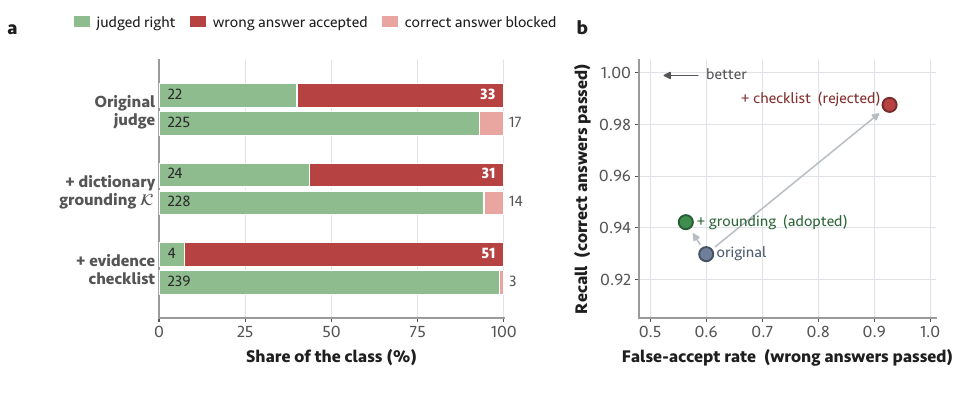}
\caption{RQ1: Comparison of Judge designs on $297$ executable outputs. (a) Verdicts on the $55$ incorrect and $242$ correct outputs. (b) False-accept rate versus recall; lower false acceptance and higher recall are preferable.}
\label{fig:judge}
\end{figure}

\noindent\textbf{RQ1: Detecting incorrect answers.}
Table~\ref{tab:rq2} and Figure~\ref{fig:judge} evaluate the Judge on the $297$ executable core outputs: $242$ correct and $55$ incorrect. The original Judge accepts $225$ correct outputs, giving a recall of $0.930$, but also accepts $33$ incorrect outputs, giving a false-accept rate of $0.600$. Consequently, most executable errors do not trigger revision. The deterministic heuristics do not resolve this problem: their vetoes occur on outputs that can legitimately be empty or scalar, illustrating the limitations of using result shape as a proxy for correctness.

We then ground the Judge in the same knowledge base used for generation, supplying definitions for columns used in the query and related alternatives. This change increases accepted correct outputs from $225$ to $228$ and reduces accepted incorrect outputs from $33$ to $31$, with negligible added latency in the recorded runs. One detected error involves a motorcycle question answered using \code{BIKEFLAG}, whose definition identifies bicycle involvement. The improvement is modest: the grounded Judge still accepts $56.4\%$ of incorrect outputs. An alternative evidence-linked checklist performs worse, raising the false-accept rate to $0.927$. This result shows that a more elaborate assessment prompt does not necessarily produce better discrimination; the experiment does not establish why the checklist encourages acceptance.

\begin{table}[pos=t]
\caption{RQ1: Judge verdicts against execution-match labels for the same $297$ executable core outputs. TP and FP are \textsc{pass} verdicts on correct and incorrect outputs; FN and TN are \textsc{fail} verdicts on those outputs. False-accept rate is FP/(FP+TN).}
\label{tab:rq2}
\centering\small
\setlength{\tabcolsep}{6pt}
\begin{tabular}{@{}lrrrrrrr@{}}
\toprule
\textbf{Judge design} & \textbf{TP} & \textbf{FP} & \textbf{FN} & \textbf{TN} & \textbf{Precision} & \textbf{Recall} & \textbf{False-accept} \\
\midrule
Original Judge
& 225 & 33 & 17 & 22 & $0.872$ & $0.930$ & \bad{$0.600$} \\
Dictionary-grounded Judge
& \good{228} & \good{31} & \good{14} & \good{24} & \good{$0.880$} & \good{$0.942$} & \good{$0.564$} \\
Evidence-checklist variant
& 239 & 51 & 3 & 4 & $0.824$ & $0.988$ & \bad{$0.927$} \\
\bottomrule
\end{tabular}
\end{table}

\begin{figure}[pos=t]
\centering
\includegraphics[width=\linewidth]{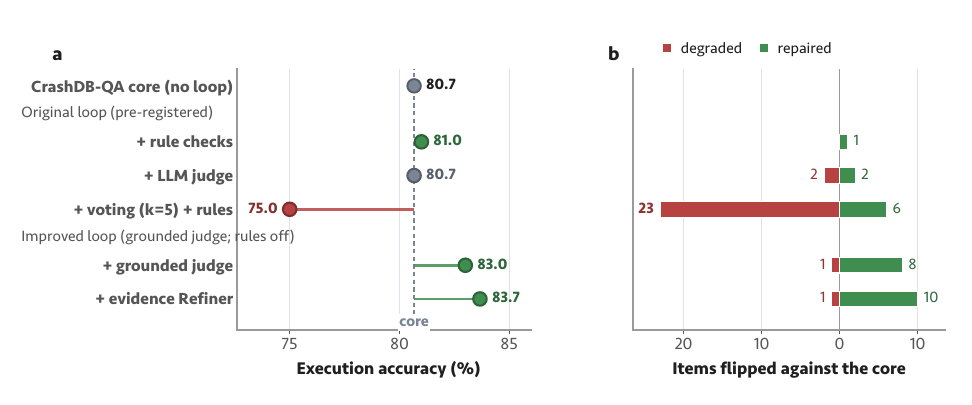}
\caption{RQ2: Effects of verification and refinement. (a) Final execution accuracy, with the core shown as a dashed reference line. (b) Repairs and regressions relative to the core.}
\label{fig:loop}
\end{figure}

\noindent\textbf{RQ2: Pre-registered verification configurations.}
The original verification configurations do not significantly improve the core, as shown in Table~\ref{tab:refiner} and Figure~\ref{fig:loop}. Rule checks increase accuracy from $80.7\%$ to $81.0\%$, repairing one answer without degrading another. The ungrounded LLM Judge produces two repairs and two regressions, leaving accuracy unchanged. Both comparisons have exact McNemar $p=1$. Combining the ungrounded Judge, rule checks, and five-candidate voting reduces accuracy to $75.0\%$: six answers are repaired, but $23$ previously correct answers become incorrect. This decline is significant, with exact $p=0.002$ and Holm-adjusted $p=0.007$, and median latency increases to nearly four times that of the core.

The replay analysis and execution traces identify two limitations of the original loop. First, the Judge accepts many incorrect answers, preventing them from receiving corrective feedback. Second, the critique-only Refiner supplies the failed SQL and critique without the returned evidence, limiting the Generator's ability to diagnose the previous result. Together, these observations motivate changes to error detection and feedback construction. The decline in the voting configuration also shows that additional candidates and verification steps can increase regressions rather than provide a monotonic improvement.

\noindent\textbf{RQ2: Grounded verification and evidence-carrying refinement.}
The diagnostic study introduces the two changes sequentially while holding the remaining settings fixed: one greedy candidate, deterministic heuristics disabled, and $N=3$. Replacing the ungrounded Judge with the grounded design raises final accuracy to $83.0\%$, with eight repairs and one regression relative to the core. The exact McNemar test gives $p=0.039$. The repairs include two motorcycle queries redirected to \code{CYCLFLAG} from bicycle and commercial-vehicle flags. Adding the evidence-carrying Refiner raises accuracy further to $83.7\%$, with ten repairs and one regression relative to the core, giving $p=0.012$. Additional repairs include breakdown queries whose returned structure exposes the aggregation error. Both configurations remain significant against the core after Holm correction within this improvement family, with adjusted $p$ values of $0.039$ and $0.023$, respectively. These tests establish improvements over the core; they do not establish that the additional gain from $83.0\%$ to $83.7\%$ is itself significant.

\begin{table}[pos=t]
\caption{RQ2: Verification configurations compared with the CrashDB-QA core on $300$ questions. All loop configurations use $N=3$ and one greedy candidate unless stated otherwise. Repairs and regressions are counted relative to the core; $p$ denotes the unadjusted exact McNemar test.}
\label{tab:refiner}
\centering\small
\setlength{\tabcolsep}{3.5pt}
\begin{tabular}{@{}lllrccc@{}}
\toprule
\textbf{Configuration} & \textbf{Judge} & \textbf{Feedback} & \textbf{Acc.\ (\%)} & \textbf{Repairs} & \textbf{Regressions} & $p$ \\
\midrule
Core only
& --- & --- & $80.7$ & --- & --- & --- \\
\midrule
\tabgroup{7}{Pre-registered configurations}
Rule checks
& rules only & critique & $81.0$ & $1$ & $0$ & $1$ \\
LLM Judge
& LLM & critique & $80.7$ & $2$ & $2$ & $1$ \\
Voting, $k{=}5$, + rules
& LLM + rules & critique & \bad{$75.0$} & $6$ & \bad{$23$} & $0.002$ \\
\midrule
\tabgroup{7}{Diagnostic improvements: deterministic heuristics disabled}
Grounded Judge
& LLM + $\K$ & critique & $83.0$ & $8$ & $1$ & $0.039$ \\
+ evidence Refiner
& LLM + $\K$ & critique + evidence & \good{$83.7$} & \good{$10$} & $1$ & \good{$0.012$} \\
\bottomrule
\end{tabular}
\end{table}

\noindent\textbf{Scope and implications.}
In the evidence-carrying configuration, $43$ of the $300$ questions undergo additional iterations, and median end-to-end latency increases from $9.2$~s to $13.1$~s. The ten repairs span breakdowns, rankings, trends, location summaries, seasonality, and severity weighting. No C9 comparison question is repaired, and the single regression occurs in that category. The gains therefore concern errors that the current Judge can identify and the Generator can revise from the supplied feedback; they do not resolve the benchmark's most persistent compositional weakness. Moreover, the improved designs were developed after inspecting failures and evaluated on the same benchmark. Their gains provide diagnostic evidence for grounded, execution-informed repair, with independent confirmation across additional models, runs, and held-out questions remaining necessary.

Taken together, the experiments identify two distinct requirements for reliable local analysis. The core benefits from a model that combines usable SQL output with accurate analytical composition. Verification adds value when its decisions use documented field meanings and its feedback exposes the execution evidence needed for repair. Neither additional reasoning time nor a more elaborate loop consistently improves performance on its own.

\FloatBarrier
\section{Conclusion}\label{sec:conclusion}

This study presented CrashDB-QA, a fully local, schema-grounded agentic workflow that connects natural-language safety questions to database queries, execution results, and plain-language explanations. The workflow combines specialized language-model tasks with deterministic controls for schema validation and read-only execution, while retaining intermediate outputs for inspection. Its optional extension, CrashDB-QA+, uses documented field definitions and execution evidence to support verification and repair. We also introduced a $300$-question benchmark spanning ten analytical categories, providing an execution-based assessment of both routine retrieval and more demanding analytical composition.

Evaluation across nine local open-weight models shows that practical performance depends on executable output, domain interpretation, and analytical composition. Qwen-2.5-Coder 7B achieved the highest core accuracy of $80.7\%$ at a median latency of $9.2$~s, providing a favorable accuracy--latency balance in the evaluated environment. Retrieval tasks were generally more reliable than reasoning tasks, and comparisons between groups or periods remained difficult across all models. Verification produced mixed results: the original configurations offered no significant improvement, and the combination of voting and verification reduced accuracy. Grounding the Judge and supplying execution evidence to the Refiner increased accuracy to $83.7\%$, with ten repairs and one regression relative to the core, at a median latency of $13.1$~s. These findings indicate that effective refinement depends on the quality of error detection and corrective evidence rather than the number of verification steps.

The broader contribution is an auditable approach to accessing structured safety information. By linking an analyst's request to documented definitions, generated SQL, and returned values, CrashDB-QA makes the basis of an answer available for review. The benchmark further shows why reliability should be assessed by analytical task: strong performance on counts and summaries does not establish competence in comparisons or derived measures. This distinction can help agencies identify suitable uses and determine where expert review remains necessary. The same design principles may also support other transportation databases with documented schemas and local deployment requirements.

Several limitations define the scope of these findings. Evaluation used one statewide schema and a single frozen snapshot, and execution agreement cannot establish semantic equivalence across all possible database instances. The verification improvements were developed through failure analysis and tested on the same benchmark using one model, so independent confirmation is needed. The Judge still accepted many incorrect outputs, and the case studies showed that explanation errors can arise even after query execution. Future work will extend evaluation across held-out questions, database schemas, and runtime configurations; develop stronger methods for comparative analysis and unsupported-request detection; and assess explanation faithfulness. Multi-table analysis and field studies of analyst accuracy, review effort, and time savings will further establish the workflow's practical value.

\section*{Acknowledgments}
The Community Maps Predictive Analytics Tool and WisTransPortal MV4000 Crash Database were developed through sponsorship and collaboration with the Wisconsin Department of Transportation. 




\bibliographystyle{model1-num-names}
\bibliography{references}

\end{document}